\documentclass{bmvc2k}
\usepackage{amsmath}
\usepackage{algorithm}  
\usepackage{algorithmic}
\usepackage{amssymb}

\title{Modeling Explicit Concerning States for Reinforcement Learning in Visual Dialogue}

\addauthor{Zipeng Xu}{xuzp@bupt.edu.cn}{1}
\addauthor{Fandong Meng}{fandongmeng@tencent.com}{2}
\addauthor{Xiaojie Wang}{xjwang@bupt.edu.cn}{1}
\addauthor{Duo Zheng}{zd@bupt.edu.cn}{1}
\addauthor{Chenxu Lv}{chenxulv@bupt.edu.cn}{1}
\addauthor{Jie Zhou}{withtomzhou@tencent.com}{2}
\addinstitution{
 Beijing University of Posts and Telecommunications,\\
 Beijing, China
}
\addinstitution{
 Pattern Recognition Center, \\
 WeChat AI, \\
 Tencent Inc., China
}

\runninghead{Z. Xu, F. Meng, X. Wang, ET AL.}{Modeling Explicit Concerning States}

\def\etal{\emph{et al}\bmvaOneDot}

\begin{document}

\maketitle

\begin{abstract}
To encourage AI agents to conduct meaningful Visual Dialogue (VD), the use of Reinforcement Learning has been proven potential. In Reinforcement Learning, it is crucial to represent states and assign rewards based on the action-caused transitions of states. However, the state representation in previous Visual Dialogue works uses the textual information only and its transitions are implicit. In this paper, we propose Explicit Concerning States (ECS) to represent what visual contents are concerned at each round and what have been concerned throughout the Visual Dialogue. ECS is modeled from multimodal information and is represented explicitly. Based on ECS, we formulate two intuitive and interpretable rewards to encourage the Visual Dialogue agents to converse on diverse and informative visual information. Experimental results on the VisDial v1.0 dataset show our method enables the Visual Dialogue agents to generate more visual coherent, less repetitive and more visual informative dialogues compared with previous methods, according to multiple automatic metrics, human study and qualitative analysis.\footnote {The code is available at \url{https://github.com/zipengxuc/ecs-visdial-rl}.}
\end{abstract}

\section{Introduction}
\label{sec:intro}

Visual Dialogue (VD) aims to create intelligent conversational systems that understand vision and language.
It is one of the ultimate goals in Artificial Intelligence (AI) \cite{winograd1972understanding}.
Although important, it is still challenging to generate coherent, visual-related and informative utterances based on the complete understanding of vision and dialog history.

To approach the target, Das \etal~\cite{das2017learning} propose a Q-Bot-A-Bot image-guessing game, where Q-Bot (who cannot see the image) raises question at each round and A-Bot (who can see the image) answers, and introduce the use of Reinforcement Learning (RL) to optimize the two Visual Dialogue agents goal-orientedly. 
In RL, a suitable state representation is regarded as a fundamental part in the learning process \cite{jones2010integrating} and has long been studied in textual dialogue tasks \cite{young2010hidden, williams2014dialog, henderson2014word, mrksic-etal-2017-neural, wu-etal-2019-transferable}.
Similarly in Visual Dialogue, it is vital that the state representation can be able to help distinguish whether the currently generated utterances bring about new useful visual information.
Consequently, the reward, which is formulated from the transitions of states, can constantly encourage the agents to converse on new visual informative utterances and achieve success in image-guessing.

However, previous works \cite{das2017learning, DBLP:journals/corr/abs-1808-04359, murahari2019visdialdiversity} represent the state in Visual Dialogue in an insufficient and implicit way, resulting in ineffective rewards.
As they derive the state representation in Visual Dialogue from the hidden state of the RNN-encoded textual dialogue history, there are several disadvantages: 1) it is insufficient to represent the Visual Dialogue state using textual information merely; 2) the transitions of states are implicit for it can only be measured between successive hidden states; 3) as dialogue history is comparatively long, the change of hidden state before and after the encoding of new question-answer pair can be subtle.
Since the rewards judge whether the new utterances are informative based on the distances between the state representation and target image at successive rounds, thus its effects diminish as dialogue progresses. 
Experimental results show that the agents always repeat themselves, and the image-guessing performance stops increasing at an early stage, indicating the rewards are unable to constantly encourage agents to generate new visual informative utterances. 
Vishvak \etal \cite{murahari2019visdialdiversity} try to fix dialogue repetition by penalizing the similarity of dialogue hidden states between successive rounds. 
Yet, this method cannot technically avoid repetition between every alternate dialogue round (at t and t+2). 
Besides, the auxiliary objective built from single modality is inadequate for the multi-modal task.
\begin{figure}
  \centering
  \includegraphics[width=0.95\linewidth]{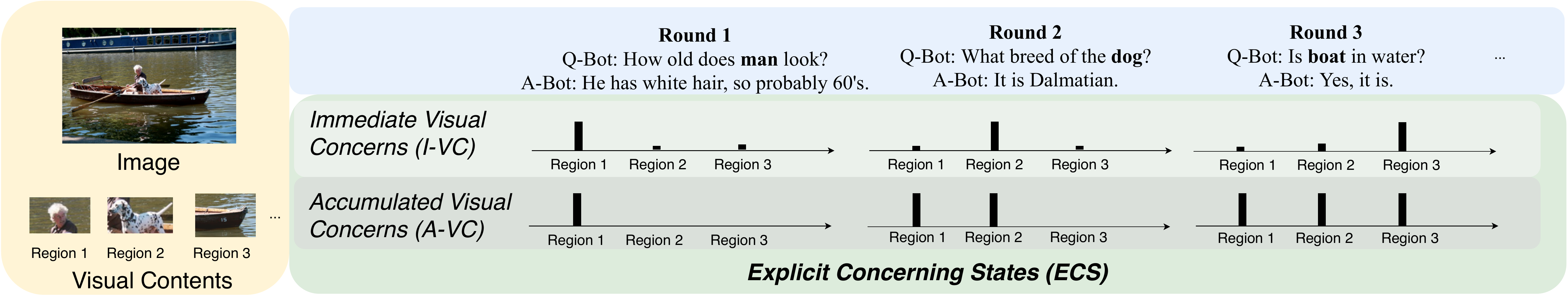}
  \caption{Illustration of Explicit Concerning State (ECS) in the Visual Dialogue process.}
 \label{fig1}
\end{figure}

To remedy the problems, in this paper, we propose Explicit Concerning States (ECS) to represent the state in Visual Dialogue efficiently and explicitly. 
ECS is a 2-tuple: <Immediate Visual Concerns (I-VC), Accumulated Visual Concerns (A-VC)>. 
As in Fig.~\ref{fig1}, I-VC models the visual contents that Q-Bot concerns immediately, A-VC models the visual contents that Q-Bot has concerned throughout the dialogue.
I-VC is formalized as the question-history guided visual attention distribution on target image. 
A-VC is the accumulated binary states, representing whether the corresponding contents have been covered or not.
Contrary to previous methods, the advantages of ECS are as follows: 1) ECS is efficiently established based on both visual and textual information; 2) the changing Visual Dialogue states are explicitly related to the visual contents in image; 3) whether Q-Bot refers new visual contents related to the target image can be clearly measured by A-VC.

Based on ECS, we further propose two intuitive and interpretable rewards to encourage the agents to constantly generate new, visual related and informative utterances.
Diversity Reward encourages the diversity of I-VC at successive rounds so as to encourage Q-Bot to ask different visual information. 
Informativity Reward encourages the growth of A-VC so as to encourage Q-Bot to constantly ask new visual information as dialogue progresses. 
Meanwhile, A-Bot is also improved as correct and detailed response to current question will guide Q-Bot to transfer its attention to new visual contents.

We evaluate our approach on the large-scale VisDial v1.0 dataset \cite{das2017visual}. 
Experimental results demonstrate the effectiveness and superiority of our approach compared with previous methods, according to multiple automatic metrics, human study and qualitative analysis. 
To conclude, our main contributions are as follows:
\begin{itemize}
    \setlength{\itemsep}{0pt}
    \setlength{\parsep}{0pt}
    \setlength{\parskip}{0pt}
    \item We propose Explicit Concerning States (ECS) to explicitly model what is currently concerned and what has been concerned in the Visual Dialogue process.
    \item We propose two intuitive and interpretable rewards based on Explicit Concerning States (ECS) to award the agents for they generate diverse and new visual informative utterances in Reinforcement Learning.
    \item We conduct experiments on the large-scale VisDial v1.0 dataset and achieve not only improved image-guessing performance, but also less repetitive and more diverse Q-Bot, descriptive and detailed A-Bot. Human study shows our agents generate more coherent, visual-related and informative dialogues that achieve superior image retrieval results as good as human dialogues. 
\end{itemize}

\section{Background}
To improve Visual Dialog agents with Reinforcement Learning, previous works introduce a Q-Bot-A-Bot image-guessing game setting. Given an undisclosed image $I$, Q-Bot, who only knows the image caption $c$ at first, has to raise a series of questions about the undisclosed image and make guesses. A-Bot, who can see the image, gives its answer accordingly.

Q-Bot consists of a encoder to encode the textual dialogue history $H_t = \{c, (q_1, a_1), \ldots, \\(q_{t-1}, a_{t-1})\}$, a decoder to generate question $q_t$ and a feature regression network to make image prediction $\bf\hat{y}_{t}$. A-Bot consists of a multi-modal encoder, which encodes the image, dialogue history and question, and a decoder, to generate answer $a_t$.

Under the setting, RL is used to jointly optimize Q-Bot and A-Bot by assigning rewards based on the action-caused transitions of state.
At round $t$, given dialogue history $H_t = \{c, (q_1, a_1), \ldots, (q_{t-1}, a_{t-1})\}$, Q-Bot generates $q_t$ and A-Bot generates $a_t$. 
In previous methods \cite{das2017learning, DBLP:journals/corr/abs-1808-04359, murahari2019visdialdiversity}, w.r.t $S_t = \{I, H_t\}$ and action $A_t$ (i.e., selecting the words from the vocabulary $V$ to generate $q_t$ and $a_t$), the reward is formulated as:
\begin{equation}
r(S_t, A_t) = ||{\bf y^{gt}} - {\bf \hat{y}_{t-1}}||_2^2 - ||{\bf y^{gt}} - {\bf\hat{y}_{t}}||_2^2,
\label{eq:1}
\end{equation}
where $\bf y^{gt}$ is the image feature of target $I$ and $\bf\hat{y}_{t}$ is the predicted image feature, which is derived from Q-Bot by passing the hidden state of the RNN-encoded textual dialogue through the feature regression network, i.e., a fully-connected layer.
REINFORCE algorithm \cite{williams1992simple} is used to update agents' parameters with the designed reward.

\section{Explicit Concerning States in VD}
\begin{figure}
  \centering
  \includegraphics[width=0.9\linewidth]{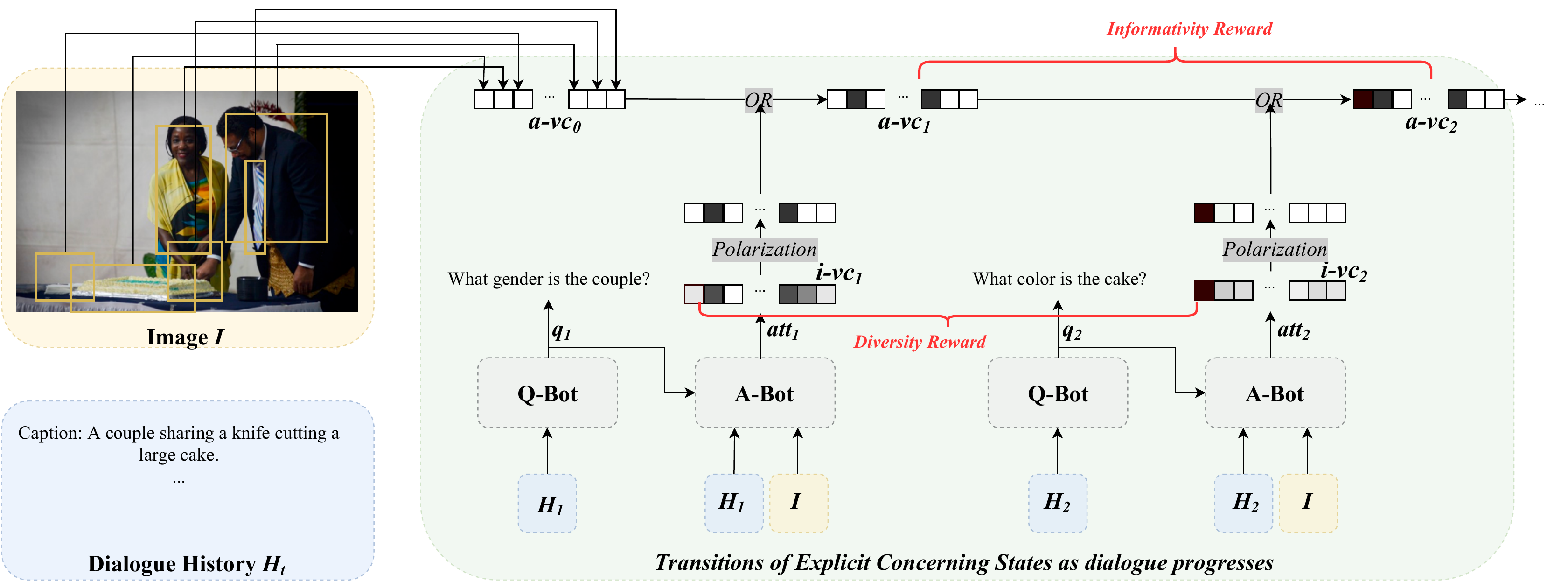}
  \caption{Overview of the proposed ECS and ECS-based reward formulations.}
  \label{fig2}
\end{figure}

Explicit Concerning States (ECS) is a 2-tuple: <Immediate Visual Concerns (I-VC), Accumulated Visual Concerns (A-VC)>. As in Fig.~\ref{fig2}, I-VC models what visual contents that Q-Bot concerns immediately, and A-VC models what visual contents that Q-Bot has concerned throughout the dialogue. As Q-Bot and A-Bot are jointly optimized in a cooperative setting, ECS is modeled from A-Bot's view, concretely based on some existed computation in an attention-based A-Bot.

Firstly, we propose I-VC to model the immediately concerned visual contents. Considering the co-reference phenomenon, we formalize I-VC based on the question-history-guided attention on image in an attention-based A-Bot. As the common visual attention mechanisms in A-Bot, at round $t$, current question correlated visual attention is guided by current question $q_t$ and history $H_t$. Accordingly, I-VC at round $t$ is formulated as:
\begin{equation}
\label{eq:2}
{\bf {i\text{-}vc}_t} = {\bf att_t} = P_{A-Bot}(o|I, H_{t}, q_t),
\end{equation}
where $\emph{o}$ is the set of all objects or spatial grids in $I$ and ${i\text{-}vc}_t^j$ is the attention weight for $o_j \in \emph{o}$.

Further, we propose A-VC to model the accumulated concerned visual contents, i.e., the visual contents that have been concerned by Q-Bot throughout the dialogue. A-VC is formulated as the accumulated binary states correspond with $\{o_j\}_{j=1}^N$, where $N$ is the number of all object proposals or spatial grids. At t-th round, ${a\text{-}vc}_t^j = 0$ means the $o_j$ has not been referred in the past t-round dialogue while ${a\text{-}vc}_t^j = 1$ otherwise.
In concrete, A-VC is initialized with $\textbf{0} \in R^N$, then is updated by two steps as dialogue progresses:

Step 1, compute the polarized I-VC by projecting the attention weights in I-VC into binary states by the following operations:
\begin{equation}
norm({\bf {i\text{-}vc}_t}) = \frac{{\bf {i\text{-}vc}_t} - min({\bf {i\text{-}vc}_t})}{max({\bf {i\text{-}vc}_t})-min({\bf {i\text{-}vc}_t})},
\end{equation}
\begin{equation}
\label{eq:polar}
polarize({i\text{-}vc}_t^j) = \left\{
\begin{aligned}
1 & , & if\;norm({i\text{-}vc}_t^j) > \gamma,  \\
0 & , & else.
\end{aligned}
\right.
\end{equation}
Accordingly, ${i\text{-}vc}_t^j \in (0,1)$ is projected into a binary value in \{0, 1\} with a threshold $\gamma$. This is conducted to have a definite identification of each visual content $o_j \in \emph{o}$ is immediately concerned by Q-Bot or not: the $polarize({i\text{-}vc}_t^j)=1$ means Q-Bot's immediately concerned visual contents include $o_j$ while $polarize({i\text{-}vc}_t^j)=0$ otherwise.

Step 2, update A-VC with the polarized I-VC by the logic $OR$ operation:
\begin{equation}
{\bf {a\text{-}vc}_t} = OR({\bf{a\text{-}vc}_{t-1}}, polarize({\bf{i\text{-}vc}_t})) 
\end{equation}
At each round, A-VC is dynamically updated. Only when current question concerns new visual contents (e.g., $o_j$, whose ${a\text{-}vc}_{t-1}^j$ is $0$ and $polarize({i\text{-}vc}_t^j)$ is $1$), will it lead to the growth of ${\bf {a\text{-}vc}_t}$, because ${a\text{-}vc}_t^j$ will turn from $0$ to $1$. Accordingly, whether Q-Bot refers to new visual contents in the image is clearly measured by comparing A-VC at different rounds. 

With I-VC and A-VC, ECS efficiently and explicitly represents the state in Visual Dialogue. Using both visual and textual information enables more accurate modeling results as compared with previous single modal methods. I-VC and A-VC thus provide reliable references for reward formulations.

\section{ECS-based RL framework}

\subsection{Learning Environment}
\label{sec:4.1}
Learning Environment includes the supervisedly pretrained Q-Bot and A-Bot. We use the same Q-Bot as in previous work \cite{das2017learning}. To establish ECS, we need an attention-based A-Bot. Among many potential models, we choose to use a classic attention-based A-Bot with History-Conditioned Image Attentive Encoder (HCIAE) \cite{lu2017best} to verify the effectiveness of our method. For image feature, we use the static features provided by bottom-up attention \cite{anderson2018bottom}, which is trained based on the framework of Faster-RCNN \cite{ren2015faster} and the annotated data in Visual Genome \cite{krishna2016visual}. The method adaptively extracts K (ranging from 10 to 100) object proposals for each image. Accordingly, $I$ is represented by $\{{\bf i_1}, {\bf i_2}, \ldots, {\bf i_K}\}$.


\subsection{Reward Formulations}
Based on ECS, we formulate two intuitive and interpretable rewards, which are Diversity Reward (DR) and Informativity Reward (IR), to award agents with reliable reasons at different Visual Dialogue states. Fig.~\ref{fig2} illustrates the two rewards in the Visual Dialogue process.

\noindent{\bf Diversity Reward (DR):} DR is to award the agents for Q-Bot holds different visual concerns at successive rounds. As I-VC represents what visual contents that Q-Bot concerns immediately, DR is formalized as the KL-divergence between successive I-VCs:
\begin{equation}
r^D(S_t, A_t) = D_{KL}({\bf {i\text{-}vc}_t}||{\bf {i\text{-}vc}_{t-1}}).
\end{equation}
On the one hand, DR encourages Q-Bot to successively ask diverse visual contents. On the other hand, it encourages A-Bot to give correct and detailed answers, as they will enable Q-Bot to transfer its attention to new visual contents.

\noindent{\bf Informativity Reward (IR):} IR is to award the agents for Q-Bot constantly concerns new visual contents in the undisclosed image as dialogue progresses. As A-VC represents what visual contents have been covered by Q-Bot throughout the dialogue, IR corresponds with constant growth of A-VC. Concretely, IR is formalized as:
\begin{equation}
r^I(S_t, A_t) = \left\{
\begin{aligned}
& 1 , & if\;\sum_{j=1}^N\;{a\text{-}vc}_t^j - {a\text{-}vc}_{t-1}^j > 0, \\
& 0 , & else.
\end{aligned}
\right.
\end{equation}
Only when Q-Bot asks about new visual contents in current round will the agents be awarded.

To conclude, DR encourages Q-Bot to question about different visual contents at successive rounds: \emph{`A'} now, \emph{`B'} next. However, DR can not avoid the repetition at alternate rounds as it will indiscriminately award \emph{`A'}-\emph{`B'}-\emph{`A'}-\emph{`B'} and \emph{`A'}-\emph{`B'}-\emph{`C'}-\emph{`D'}. To make up for this, IR is designed to only award the agents for Q-Bot questions new. Jointly using DR and IR enables 1) a Q-Bot that will not repeat itself and will keep exploring the unknown areas, and 2) an A-Bot that tends to give detailed responses so as to guide Q-Bot to transfer its attention to new contents, and consequently the visual informative dialogues.

\subsection{Training in RL}
Together with the Diversity Reward, the Informativity Reward and the original reward (annotated as $r^O$) as described in Eq.~\ref{eq:1}, the final reward is assigned as:
\begin{equation}
\label{eq:reward}
r_t = \alpha_Or_t^O + \alpha_Dr_t^D + \alpha_Ir_t^I,
\end{equation}
where $\alpha_O$, $\alpha_D$ and $\alpha_I$ are coefficients for each reward, respectively. 
With the final reward $r_t$, we use the same REINFORCE algorithm as previous methods to update policy parameters. More details about the algorithm are given in Appendix~\ref{apd:a}.

\section{Experiments}
We conduct experiments on the large-scale VisDial v1.0 dataset. The train split consists of 123k dialogues on MS-COCO \cite{lin2014microsoft} images and the validation split consists of 2k dialogues on Flickr images. All dialogues last for 10 rounds for each image. 

We compare our ECS-based rewards enhanced agents with previous state-of-the art methods. The comparing methods are: 1) \emph{`SL'}: the supervisedly pretrained basic Q-Bot and A-Bot; 2)\emph{`RL'}: basic Q-Bot and A-Bot fine-tuned by RL using the original reward \cite{das2017learning}; 3)\emph{`RL - diverse'} \cite{murahari2019visdialdiversity}: diverse Q-Bot (optimized with an additional objective) and A-Bot fine-tuned by RL using the original reward. Our full method is represented as \emph{`RL-Ours'}, i.e., basic Q-Bot and A-Bot fine-tuned by RL using the ECS-based rewards in addition. To make a fair comparison, all methods are implemented based on the same learning environment (as introduced in Sec.~\ref{sec:4.1}). Training details are given in Appendix~\ref{apd:b}.

\subsection{Results}
\begin{figure} \centering    
\subfigure[Q-Bot performance]{
\includegraphics[width=0.7\linewidth]{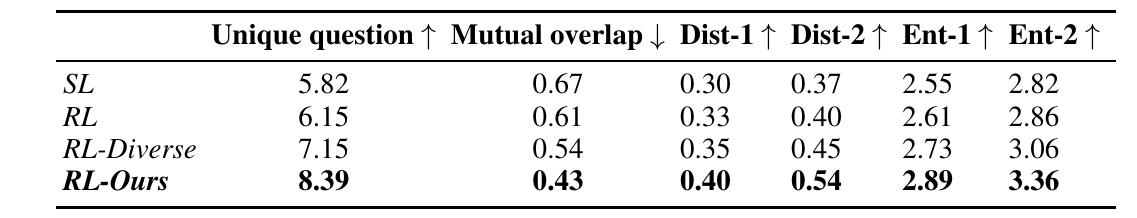}
\label{fig3:a}
}

\subfigure[A-Bot performance] {
 \label{fig:a}     
 \begin{minipage}[t]{0.5\linewidth}
 \includegraphics[width=0.95\linewidth]{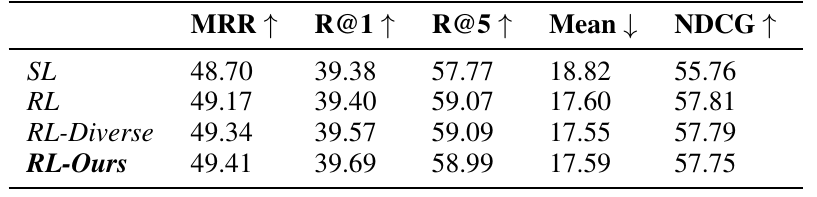}
 \label{fig3:b}
 \end{minipage}
}   
\subfigure[Image-guessing performance] { 
\label{fig:b}     
\begin{minipage}[t]{0.45\linewidth}
\includegraphics[width=0.45\linewidth]{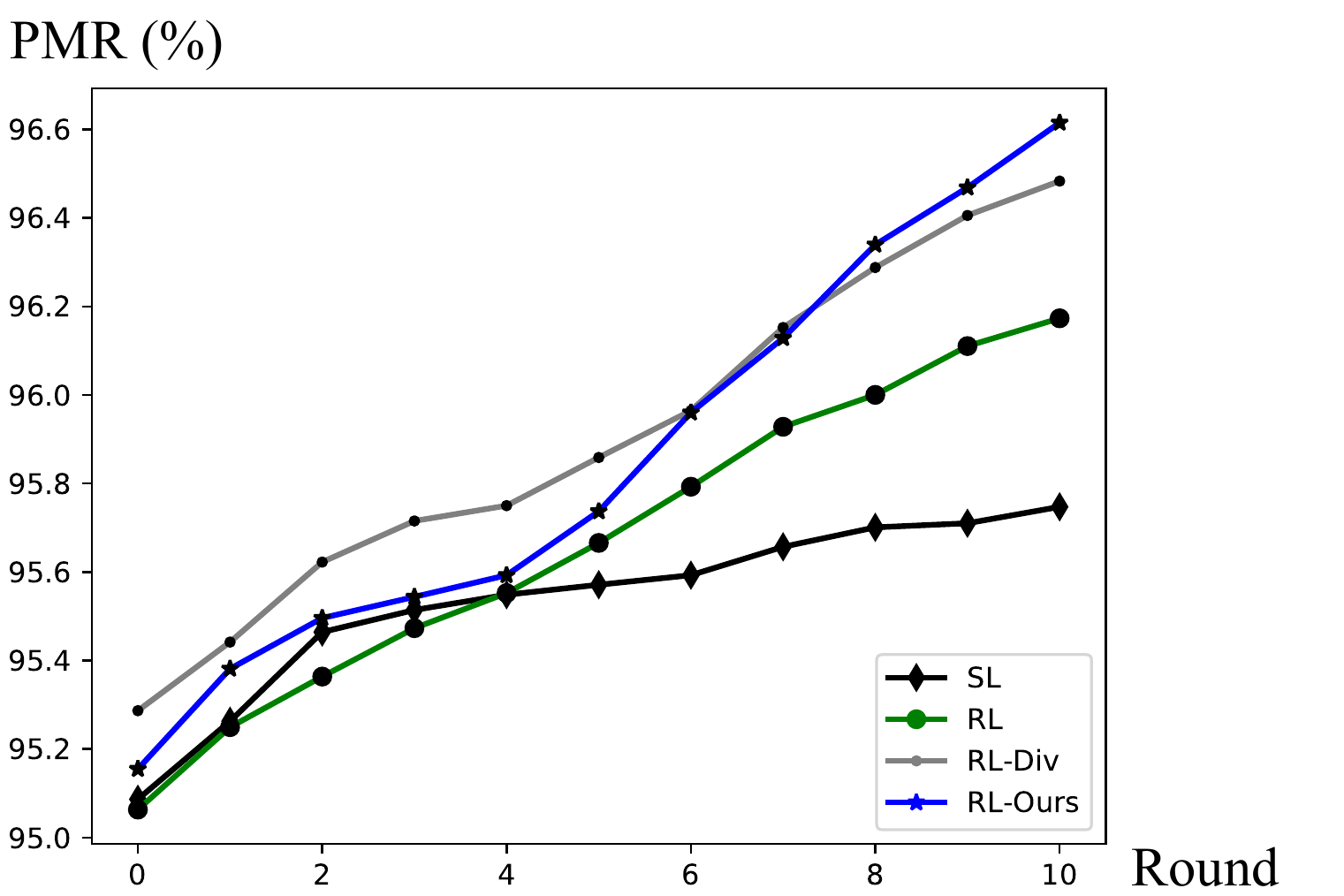}  
\includegraphics[width=0.5\linewidth]{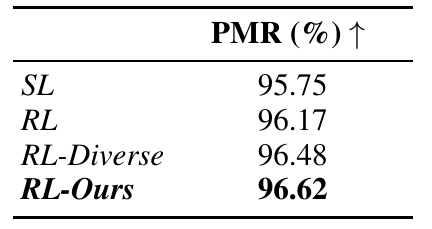}    
\label{fig3:c}
\end{minipage}
}
\caption{Quantitative results on VisDial v1.0 val.}
\label{quantR}
\end{figure}

\textbf{Q-Bot Performance.} We evaluate Q-Bot performance with the following metrics: 1) Unique Question: the number of unique questions per dialogue instance; 2) Mutual Overlap \cite{DBLP:journals/corr/abs-1805-12589}: the average value of BLEU-4 overlap of each question in the generated 10-round dialogue with the other 9 questions; 3) Dist-n and Ent-n \cite{li-etal-2016-diversity, NEURIPS2018_23ce1851}: the number and entropy of unique n-grams in the generated questions, normalized by the number of total tokens.
As shown in Fig.~\ref{fig3:a}, jointly using the ECS-based rewards significantly improves Q-Bot performance as compared to previous methods. Our method, i.e., \emph{`RL-Ours'}, generates 8.39 unique question in the 10-round dialogue in average, indicating the agent's superior ability to conduct less repetitive dialog. Besides, the results on Mutual Overlap, Dist-n and Ent-n show our method helps generate question with higher language diversity.

\noindent\textbf{A-Bot Performance.} We evaluate A-Bot's answers to human questions in VisDial v1.0 dataset with the retrieval metrics introduced in \cite{visdial}.
Fig.~\ref{fig3:b} gives the comparing results of A-Bot performance. According to the automatic evaluation metrics, the improvements brought by our method is not significant. However, our goal is to motivate the answering agents to generate more descriptive, detailed and visual informative responses which the retrieval-based metrics may not be good at measuring. In the following sections and Appendix~\ref{apd:c}, we further present the distinct characteristics of our A-Bot, which achieves positive results that meet our expectations.

\noindent\textbf{Image-Guessing Performance.} Following previous methods, we illustrate the image-guessing performance of each model in Fig.~\ref{fig3:c}. Concretely, Q-Bot and A-Bot generate 10-round dialogues via self-talk and each Q-Bot makes the image feature predictions based on the generated dialogues.  Then the candidate images in val-set are ranked and PMR (Percentile Mean Rank) is derived accordingly.
It is worth mentioning that evaluation standards vary because different prediction models are used. High PMR at this point does not guarantee the high quality of the generated dialog, but its trends indicate whether the rewards in each RL method constantly take effects as dialogue proceeds.

According to the PMR trends across dialogue rounds in the left part of Fig.~\ref{fig3:c}: in \emph{`SL'}, PMR stops increasing at an early stage. By contrast, Reinforcement Learning takes effects to incentive the growth. Compared to \emph{`RL'}, our method presents better performance as the growth on PMR are more significant and persistent, indicating the generated dialogues can constantly provide more effective information for guessing the target image. Compared to previous state-of-the-art method \emph{`RL-Diverse'}, which uses extra optimizing objective during the supervise learning stage, our method still achieves better performance.
\begin{figure}[t]
\subfigure[Image-Guessing]{
\begin{minipage}[t]{0.25\linewidth}
\includegraphics[width=\linewidth]{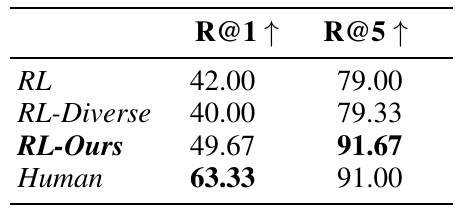}
\label{fig4:a}
\end{minipage}%
}%
\subfigure[Coherence]{
\begin{minipage}[t]{0.15\linewidth}
\centering
\includegraphics[width=\linewidth]{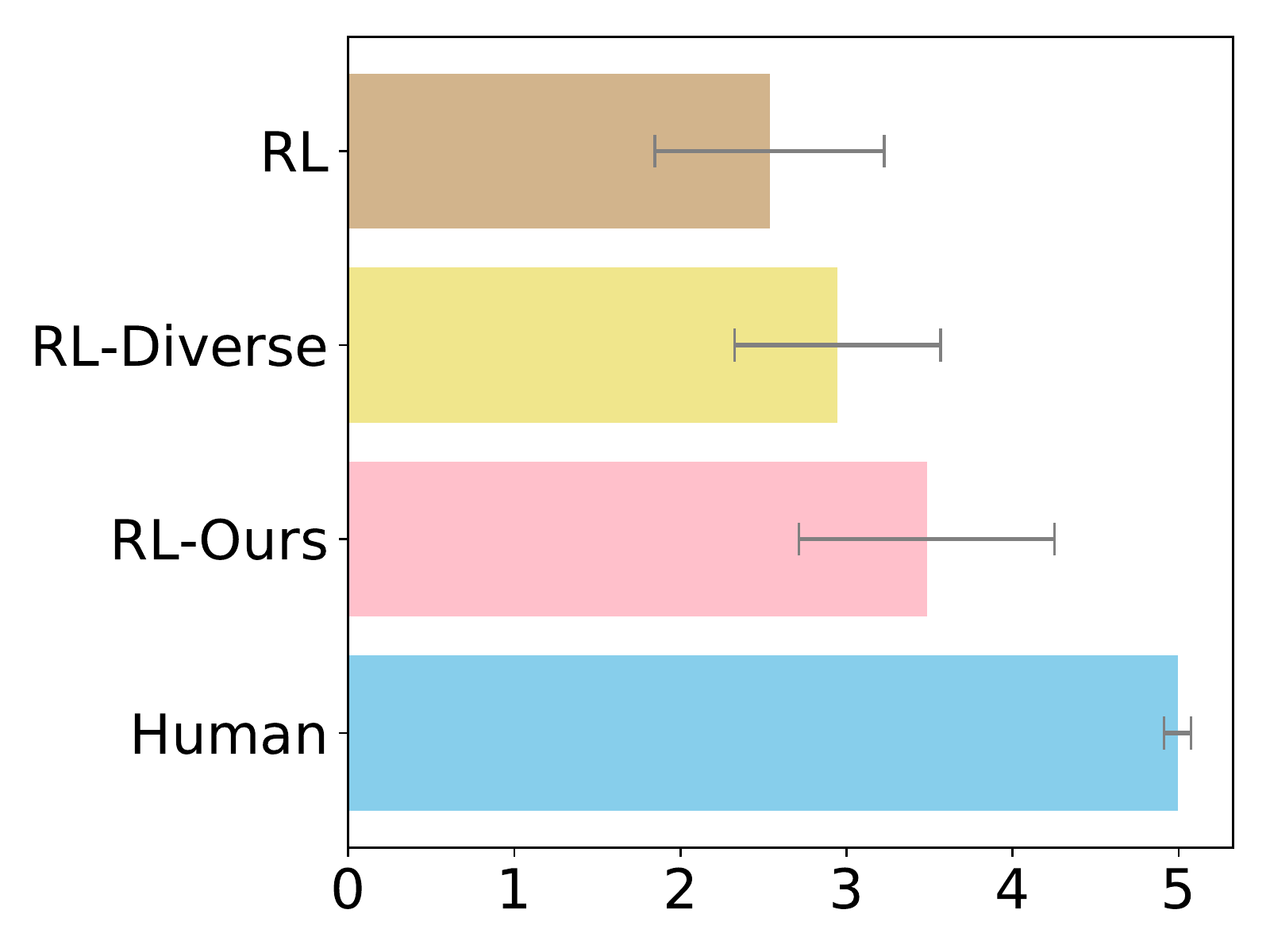}
\label{fig4:b}
\end{minipage}%
}%
\subfigure[Relevance]{
\begin{minipage}[t]{0.15\linewidth}
\includegraphics[width=\linewidth]{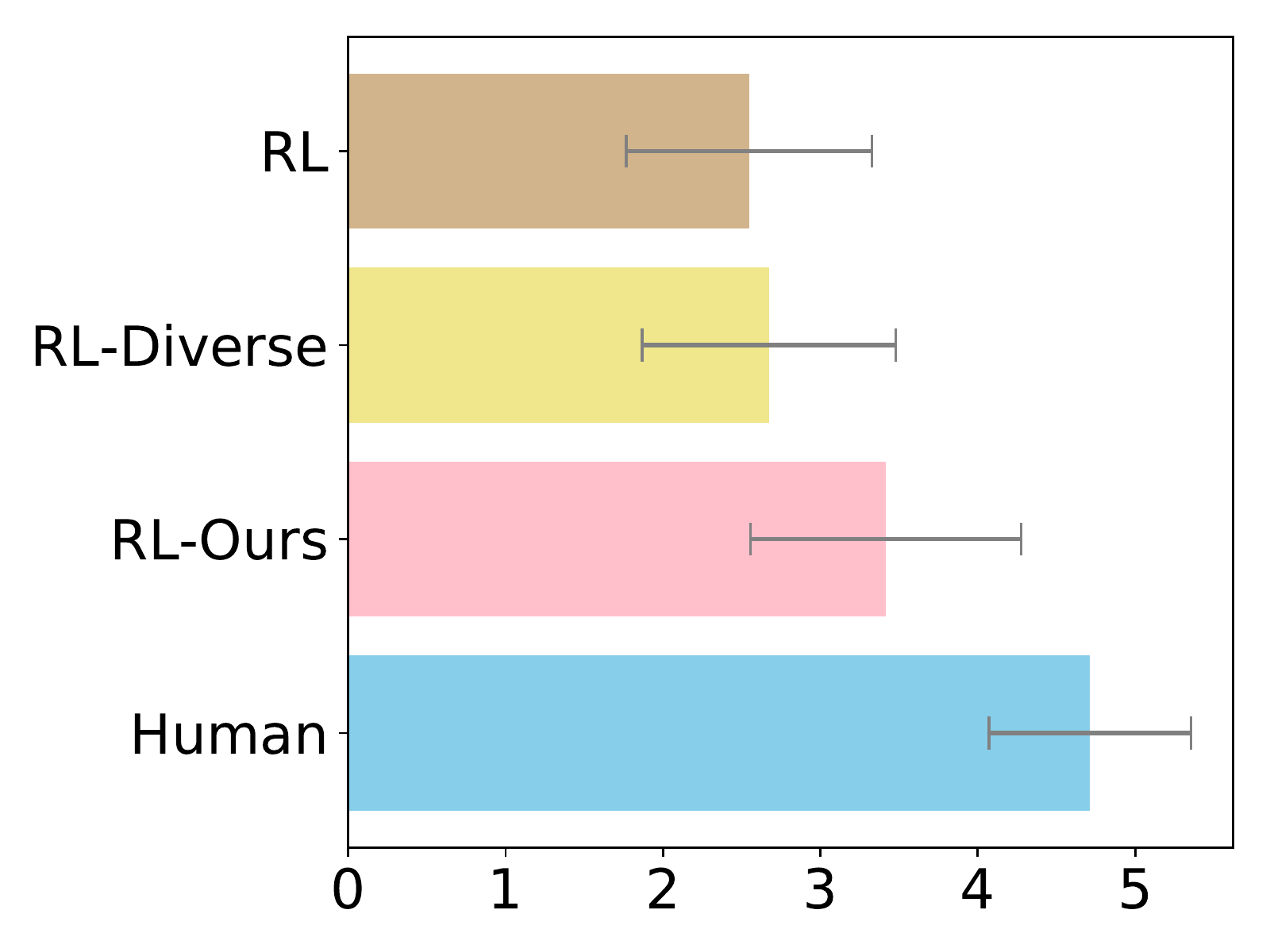}
\label{fig4:c}
\end{minipage}%
}%
\subfigure[Informativity]{
\begin{minipage}[t]{0.15\linewidth}
\includegraphics[width=\linewidth]{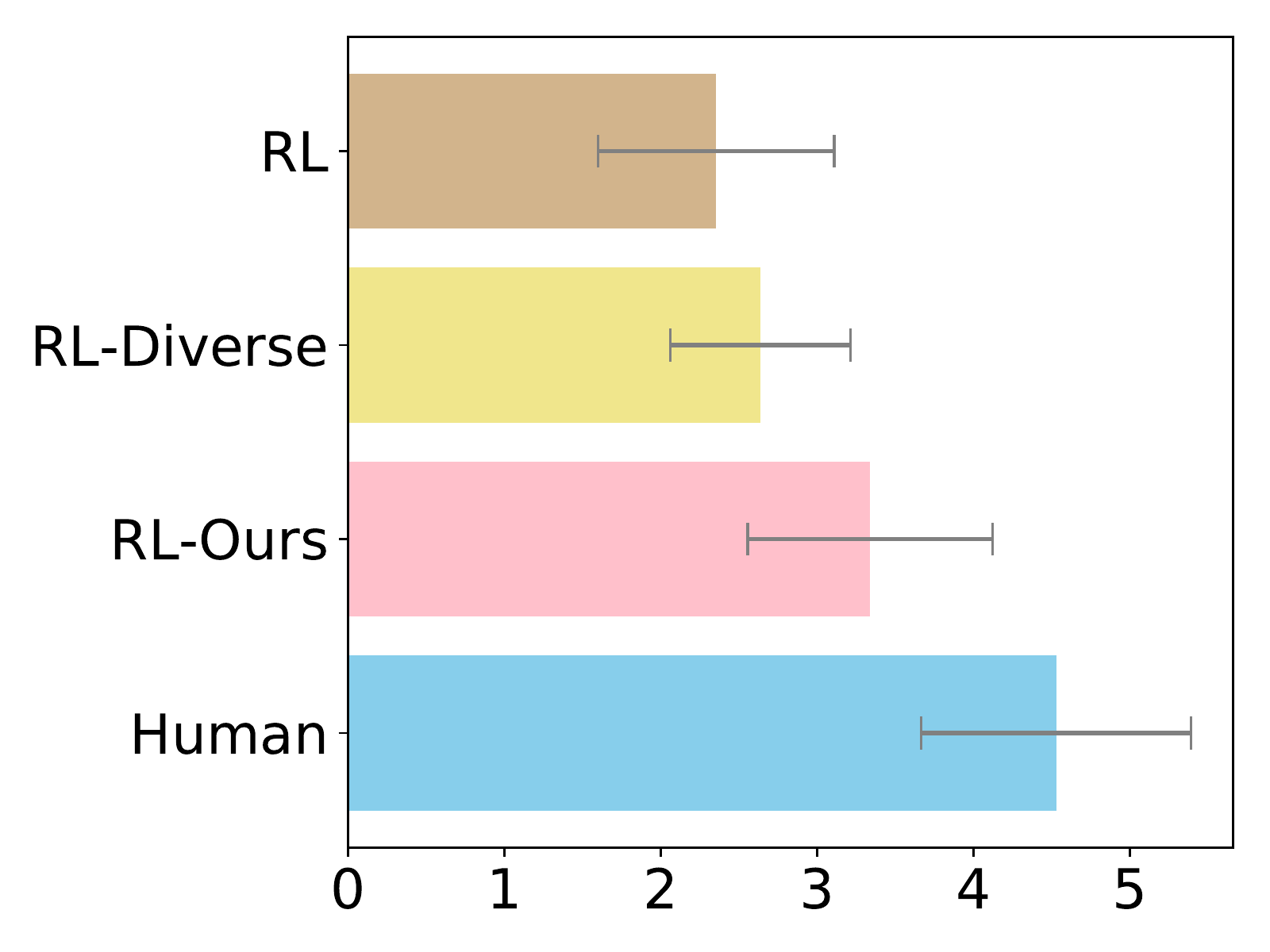}
\label{fig4:d}
\end{minipage}
}%
\subfigure[Detail]{
\begin{minipage}[t]{0.15\linewidth}
\includegraphics[width=\linewidth]{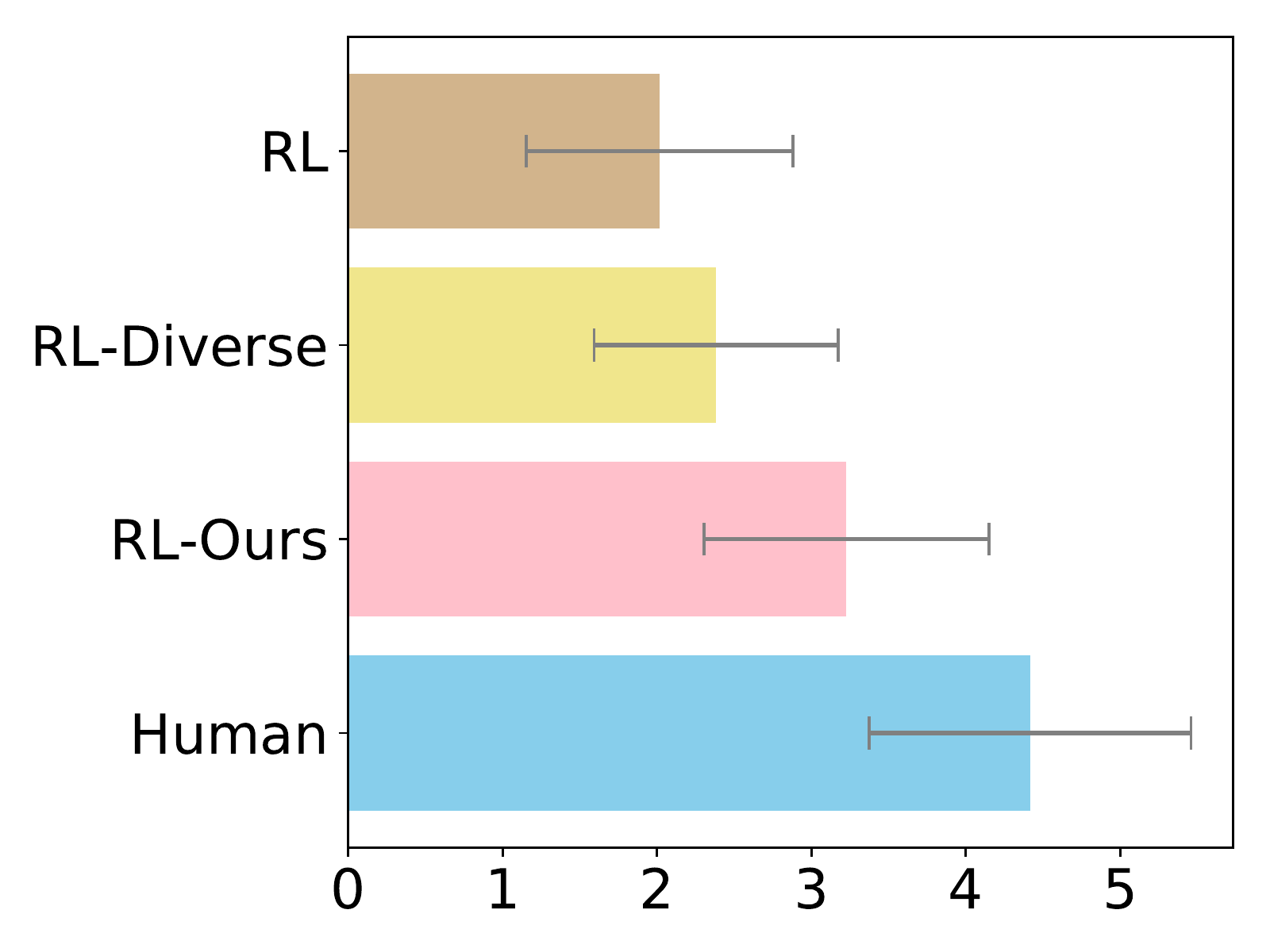}
\label{fig4:e}
\end{minipage}
}%
\centering
\caption{Results on human study.}
\end{figure}

\noindent\textbf{Human Study.} Two human evaluation tasks are conducted: dialogue-based image-guessing task and dialogue quality evaluation task. Six postgraduate students participate in and each evaluates 50 samples for each method variant.

Dialogue-based Image-Guessing task is to retrieve the target image based on the given dialogue. Following previous work \cite{murahari2019visdialdiversity}, evaluators are asked to pick top-1 and top-5 likely images from a 16-candidate-image pool and R@1 and R@5 are derived accordingly. We use the same 16-candidate-image pool which consists of the target image, 5 nearest neighbors, and 10 random images from the VisDial v1.0 test-std split. The image-guessing performances present whether the generated dialogues are human-understandable, and include enough information that helps to figure out the likely images. As shown in Fig.~\ref{fig4:a}, our method achieves best performances on R@1 and R@5 as compared to previous methods. We even outperform human on R@5, indicating that dialogues generated by our methods can cover mostly related visual contents in the target image.

Dialogue Quality Evaluation task is to rate the dialogue quality from 1 to 5 on four criteria: coherence, relevance, informativity and detail. Coherence considers whether the dialogue contexts are natural, coherent, less repetitive and consistent. Relevance assesses the degree that dialogues are relevant to the target image and given caption. Informativity defines how well the dialogues provide adequate information about the target image. Detail presents the extent that dialogues include detailed visual information. As can be seen in Fig.~\ref{fig4:b}-\ref{fig4:e}, our method achieves best scores on all criteria as compared to previous methods.

\noindent\textbf{Qualitative Analysis.} 
\begin{figure}
  \centering
  \includegraphics[width=\linewidth]{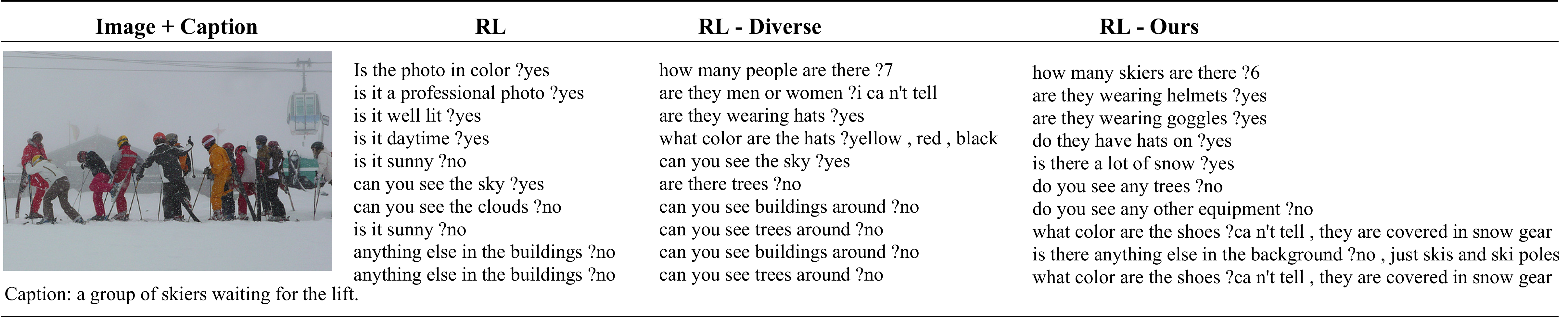}
  \caption{Generated dialogues from comparing methods.}
  \label{fig5}
\end{figure}
\begin{figure}[t]
  \centering
  \includegraphics[width=0.8\linewidth]{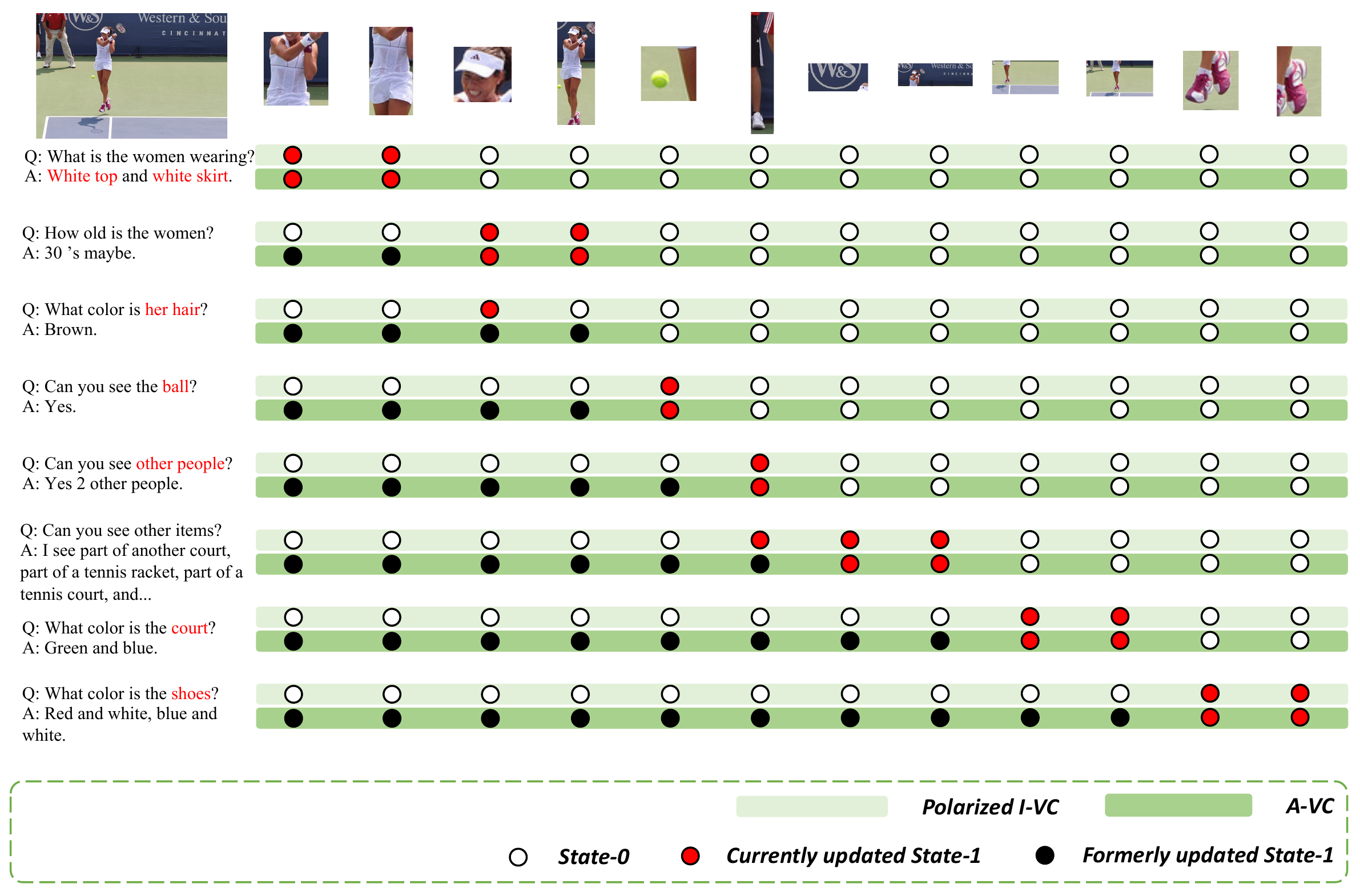}
  \caption{Visualization of ECS.}
  \label{fig9}
\end{figure}
In Fig.~\ref{fig5}, we illustrate the generated dialogue examples from comparing models. Dialogue generated by \emph{`RL'} is not very related to the main contents of the image. \emph{`RL-Diverse'} is of more diversity but always repeats at alternate turns. Besides, the description on visual contents of both models are coarse. Contrarily, our method generates dialogues that are less repetitive, more visual-coherent and more informative. As shown, dialogue generated by our methods gradually covers different contents in the image, which matches the goal of enlarging A-VC well. Besides, our dialogue agents are able to generate informative utterances with detailed visual descriptions. We are the only one to describe the \emph{``skiers"}, \emph{``wearing helmets"} and \emph{``wearing goggles"}, which are all important details of the image. Also, our A-Bot provides more descriptive and informative answers. For example, when answering \emph{``what color are the shoes''}, our agent gives additional explanations as \emph{``can't tell , they are covered in snow gear''}. Answers like this are more human-like and is helpful to keep dialog-participant interested in. More samples are given in Appendix~\ref{apd:d}. 

Besides, we visualize ECS <I-VC, A-VC> in a dialogue process in Fig.~\ref{fig9}. As dialogue progresses, Q-Bot step-wisely concerns about different visual contents related to image, raises corresponding visual-coherent questions and gradually collects detailed information of the undisclosed image. Besides, the detailed answers given by A-Bot also helps promote the dialogue process. Noticeably, when A-Bot refers \emph{``court''}, Q-Bot raises the subsequent question about the court: \emph{``What color is the court?''}. Correspondingly, the immediate concerned contents are explicitly captured by I-VC and the accumulated concerned contents are explicitly captured by A-VC. As highlighted in red, the textual contents in dialogue align well with the states on image.
During the process, I-VC varies and A-VC grows at each round. The phenomena are consistent with our design intuitions.

\subsection{Ablation Study}
\begin{table}
\scriptsize
\setlength{\tabcolsep}{0.8mm}{
\begin{center}
\begin{tabular}{l|cc|cc|c}
\hline
         & \multicolumn{2}{c|}{\textbf{Q-Bot}}  & \multicolumn{2}{c|}{\textbf{A-Bot}} &{\textbf{Image-Guessing}}  \\
        & Unique Question $\uparrow$ & Mutual Overlap $\downarrow$ & MRR $\uparrow$ & R@1 $\uparrow$  & PMR (\%) $\uparrow$  \\ 
\hline
\emph{RL}    &  6.15 & 0.61 & 49.17 & 39.40 & 96.17 \\
\emph{RL+DR}   & 7.22 & 0.53 & 49.19 & 39.42   &96.36 \\
\emph{RL+IR}    & 4.86 & 0.67 & 49.39 & 39.55 &97.35  \\
\emph{RL+DR+IR (RL-Ours)}  & 8.39 & 0.43  & 49.41 & 39.69&96.62 \\
\hline
\end{tabular}
\end{center}}
\caption{Comprehensive quantitative results on ablation study.}
\label{tab1}
\end{table}
To study the effects of individual reward, i.e., DR and IR, we conduct ablation study. Tab.~\ref{tab1} shows the comprehensive quantitative results evaluated by automatic metric on the three aspects (Q-Bot, A-Bot and image-guessing), for each ablative model.

Individually using DR (\emph{`RL+DR'}) brings about slightly better Q-Bot' diversity and image-guessing performance as compared with \emph{`RL'}, e.g. Q-Bot's Unique Question is improved from 6.15 to 7.22 and PMR is from 96.17 to 96.36.
And as IR encourages the informativity, \emph{`RL+IR'} achieves a highest PMR of 97.35, indicating the generated dialogues are more informative so as to achieve more success in guessing the target image, but the Q-Bot diversity is poor. This happens because the optimized agents are more to achieving the image-guessing goal and thus ignoring the dialogue quality. Specifically, Q-Bot learns to ask "what else?" at an early stage and A-Bot tells it nearly everything about the image.
Jointly using DR and IR achieves the most human-readable results (\emph{`RL+DR+IR'}), Q-Bot's Unique Question reaches the highest 8.39 and PMR reaches 96.62.
To conclude, DR encourages the different Q-Bot’s visual related questions at successive rounds but cannot promise the constant informativity; IR encourages the informativity of the whole dialogue but lacks guidance for generating dialogue that exchanges the information step by step; DR and IR are complementary to each other for achieving the visual coherent, less repetitive and visual informative dialogues.

\begin{figure}
  \centering
  \includegraphics[width=\linewidth]{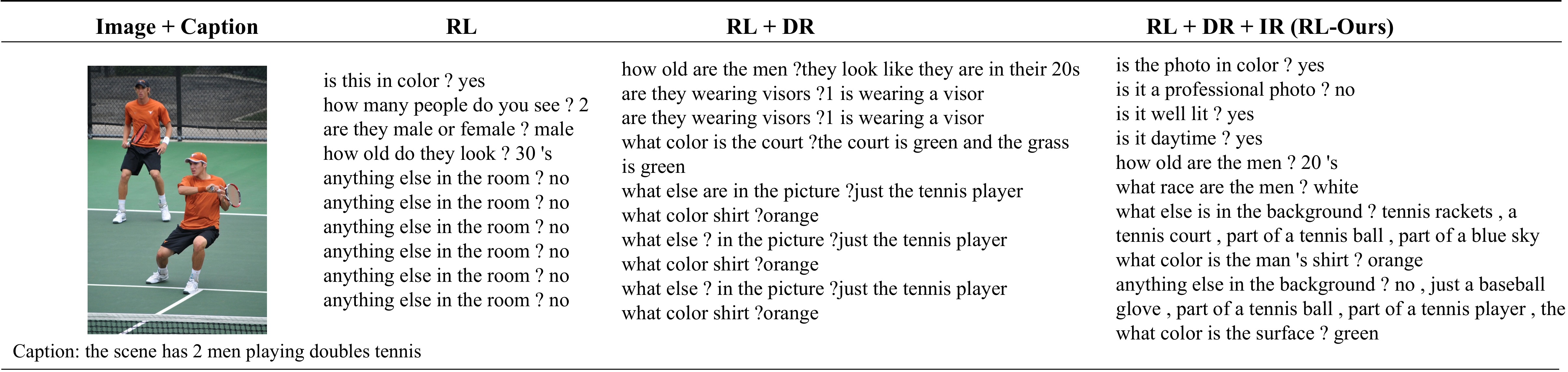}
  \caption{Qualitative results on ablation study.}
  \label{fig6}
\end{figure}

Fig.~\ref{fig6} shows the comparing qualitative results among \emph{`RL'}, \emph{`RL+DR'} and \emph{`RL+DR+IR'}. As compared to \emph{`RL'}, dialogue from \emph{`RL+DR'} converses on more visual related contents like \emph{`visors'}, \emph{`court'} and \emph{`shirt'} but suffers from repetition at successive rounds. By adding IR, the generated dialogue repeats less and is more informative.

\section{Related Work}
Our work is mostly related to previous works that improve open-domain VD using RL. Previous works \cite{das2017learning, DBLP:journals/corr/abs-1808-04359, murahari2019visdialdiversity} implicitly represent the VD state as the textual dialogue encoding, which can be insufficient for the multimodal task. To encourage diverse questions,  Vishvak \etal \cite{murahari2019visdialdiversity} propose to penalize the similarity between successive textual dialogue encodings, but the difference of textual dialogue encodings cannot promise visual informative dialogues. Contrarily, we represent the VD state in an explicit and multimodal way, enabling more reasonable and interpretable reward formulations to encourage visual informative dialogues.

Works \cite{Strub2017EndtoendOO, DBLP:journals/corr/abs-1812-06398, Abbasnejad2018WhatsTK, zhao2018learning, 10.1145/3394171.3413668} that studies using RL to improve goal-oriented VD in the Guesswhat?! task setting \cite{de2017guesswhat} are also related. Their goal is to retrieve a predefined object in the image and the answer can only be ``yes/no/not available". Designing intermediate rewards to encourage the goal-oriented visual question generation has been explored by \cite{Zhang_2018_ECCV, shukla-etal-2019-ask}, but the rewards are assigned based on external information, which needs additional annotated information and is limited to the task setting. By contrast, our reward formulations are based on internal dialogue state and are applicable in improving open-domain VD. 

Many works \cite{das2017visual, lu2017best, seo2017visual, niu2019recursive, jiang2020dam, Qi_2020_CVPR} focus on training a VD answering agent by Supervised Learning (SL). Despite high performance has been achieved according to automatic evaluation metrics, the predicted answers tend to be generic \cite{agarwal-etal-2020-history}. With the intuitive ECS-based rewards, our method facilitates an intrinsically motivated A-Bot that can generate descriptive and informative answers, and can be used to fine-tune the attention-based SL methods.

\section{Conclusion}
In this paper, we propose Explicit Concerning States (ECS) to represent the state in Visual Dialogue efficiently and explicitly, and to provide references for rewards in Reinforcement Learning. Based on ECS, we formulate Diversity Reward (DR) and Informativity Reward (IR), to award agents under different Visual Dialogue states. Boosted by the methods, our agents generate less repetitive, more visual-coherent and informative dialogues compared with previous methods according to multiple criteria.

\section{Acknowledgements}
We thank the reviewers for their comments and suggestions. This work was supported in part by the National Key Research and Development Program of China under the grant of 2020YFF0305302, and NSFC (No. 61906018), MoE-CMCC "Artificial Intelligence"  Project (No. MCM20190701).

\bibliography{main.bib}
\clearpage
\appendix
\section{Reinforcement Learning Algorithm Details}
\label{apd:a}
Based on the supervisedly pretrained Q-Bot and A-Bot, RL is used to fine-tune the two models. Following previous works, we use REINFORCE algorithm to update policy parameters. 

Q-Bot's policy is formulated as ${\pi}_Q(q_t|S_t;\theta_Q)$, where $S_t$ is the state, $q_t$ is the generated question and the policy is learned by the deep neural network parameterized by $\theta_Q$. Similarly, A-Bot's policy is formulated as ${\pi}_A(a_t|S_t;\theta_A)$. 
For policy $\pi_Q$ and $\pi_A$, the objective function of the policy gradient is given by:
$J(\theta_Q, \theta_A) = E_{\pi_Q,\pi_A}\sum_{t=1}^Tr_t(S_t, A_t)$.
Following the REINFORCE, the gradient of $J(\theta_Q)$ can be written as:
$\nabla J(\theta_Q) = E_{\pi_Q,\pi_A}r_t(\cdot)\nabla\theta_Qlog\pi_Q(q_t|S_{t})$.
Similarly, the gradient of $J(\theta_A)$ can be written as:
$\nabla J(\theta_A) = E_{\pi_Q,\pi_A}r_t(\cdot)\nabla\theta_Alog\pi_A(a_t|S_{t})$. Accordingly, Q-Bot and A-Bot are optimized with policy gradients while maximizing the reward $r_t$ as claimed in Eq.~\ref{eq:reward}.
In Algorithm~\ref{alg:1}, we exemplify the training procedure that uses RL at the beginning of the self-talking dialogue. To simplify, the target image feature is denoted by $tar\_feat$ and the predicted image feature is denoted by $pred\_feat$.

\section{Training Details}
\label{apd:b}
\noindent\textbf{Supervise Learning.} We use the same Q-Bot as in previous work \cite{das2017learning}. In concrete, the model consists of 4 components: 1) fact encoder, 2) history encoder, 3) question decoder, 4) feature regression network. Following previous settings, all encoders and decoder are 2-layer LSTM with 512-d hidden states, the feature regression network is a 1-layer MLP. Fact encoder and history encoder encodes the textual dialogue history. Based on the encoding of dialogue history, question decoder (a 2-layer LSTM) generates the question and feature regression Network  (a single fully connected layer) outputs the predicted image feature. Q-Bot is optimized with cross-entropy loss and L2 loss that minimizes the distance between predicted image feature and target image feature. Following previous works, we pretrain Q-Bot for 20 epochs with a learning rate of 1e-3 that is iteratively decayed to a minimum of 5e-5. Adam is the optimizer and dropout rate is 0.5.

\begin{algorithm}[h]
    \caption{Training the Q-Bot and A-bot using REINFORCE with ECS-based rewards.}
    \begin{algorithmic}[1]
        \FOR{Each Update}
        \STATE \textbf{\# \emph{Initialize A-VC:}}
        \STATE ${a\text{-}vc}_0 \leftarrow \bf{0}$
        \STATE ${i\text{-}vc}_0 \sim U(0,1)$
        \STATE $\triangle pred\_feat \leftarrow 0$

        \FOR{$t=1$ to $T$}
        \STATE $q_t, pred\_feat \leftarrow QBot(H_{t})$
        \STATE \textbf{\# \emph{Obtain I-VC:}}
        \STATE $a_t, att_t \leftarrow ABot(I, H_t, q_t)$
        \STATE ${i\text{-}vc}_t = att_t$
        \STATE \textbf{\# \emph{Update A-VC:}}
        \STATE ${a\text{-}vc}_t \leftarrow OR({a\text{-}vc}_{t-1}, polarize({i\text{-}vc}_t))$
        \STATE \textbf{\# \emph{Compute rewards:}}
        \STATE $r^D_t \leftarrow D_{KL}({i\text{-}vc}_t||{i\text{-}vc}_{t-1})$
        \STATE \textbf{if} $\sum_{j=1}^N\;{a\text{-}vc}_t^j - {a\text{-}vc}_{t-1}^j > 0 $ \textbf{then}
        \STATE $\;\;\;\;\;\;r^I_t \leftarrow 1$
        \STATE \textbf{else} $r^I_t \leftarrow 0$
        \STATE $r^O_t \leftarrow (tar\_feat, pred\_feat)^2 - \triangle pred\_feat$
        \STATE $r_t \leftarrow \alpha_Or_t^O + \alpha_Dr_t^D + \alpha_Ir_t^I$
        \STATE $\triangle pred\_feat \leftarrow (tar\_feat, pred\_feat)^2$

        \STATE Evaluate $\nabla J(\theta_Q)$ and update QBot
        \STATE Evaluate $\nabla J(\theta_A)$ and update ABot
        \ENDFOR
        \ENDFOR
    \end{algorithmic}
\label{alg:1}
\end{algorithm}

Among many attention-based A-Bot models, we use a classic attention-based A-Bot with History-Conditioned Image Attentive Encoder (HCIAE)~\cite{lu2017best} to verify the proposed method. According to the visual attention computation method in HCIAE, the current concerned visual contents $\bf {i\text{-}vc}_t$ in Eq.~\ref{eq:2} can further be described as:
\begin{equation}
{\bf att_t} = softmax({\bf w_a}^Ttanh({\bf W_HM_t^H} + ({\bf W_qm_t^q})\bf{1}^T),
\end{equation}
where $\bf M_t^H$ is history feature, $\bf m_t^q$ is question feature, $\bf w_a$, $\bf W_H$ and $\bf W_q$ are learnable parameters.
A-Bot is optimized with cross entropy loss. Following their implementation, we pretrain A-Bot for 40 epochs with a learning rate of 4e-4, decayed by 0.75 after 10 epochs. We use Adam as the optimizer. Dropout rate is 0.5.

\noindent\textbf{Reinforcement Learning.} The $\gamma$ in the polarization operation in Eq.~\ref{eq:polar} is set to be 0.6 empirically. The reward coefficients in Eq.~\ref{eq:reward} are set as follows: $\alpha_O$=1, $\alpha_D$=1e-1, $\alpha_I$=1e-2. The coefficients vary because different rewards are at different magnitudes. We set the coefficients based on the consideration of balancing task success and dialogue quality. We give further analysis on coefficients setting in Appendix~\ref{apd:e}.

RL smoothly begins according to a curriculum learning setting: SL is used for the first K rounds of dialogue and policy-gradient update works in the left 10 - K rounds. Following \citep{murahari2019visdialdiversity}, we start at K = 9 and gradually anneal it to 4. 
We train the agents for 30 epochs with a batch size of 32. As learning rate is inconsistent in SL, we also use different settings in RL. We optimize the Q-Bot with a learning rate of 1e-3 that is iteratively decayed up to a minimum of 5e-5 while learning rate of A-Bot is initialized with 1e-4 and decayed up to 5e-6. We clamp gradients to [-5, 5] to avoid explosion.

\section{Additional Qualitative Analysis on A-Bot}
\label{apd:c}
\begin{figure*}[t]
\centering
\subfigure[]{
\label{fig7:a}
\begin{minipage}[b]{0.9\textwidth}
\includegraphics[width=1\textwidth]{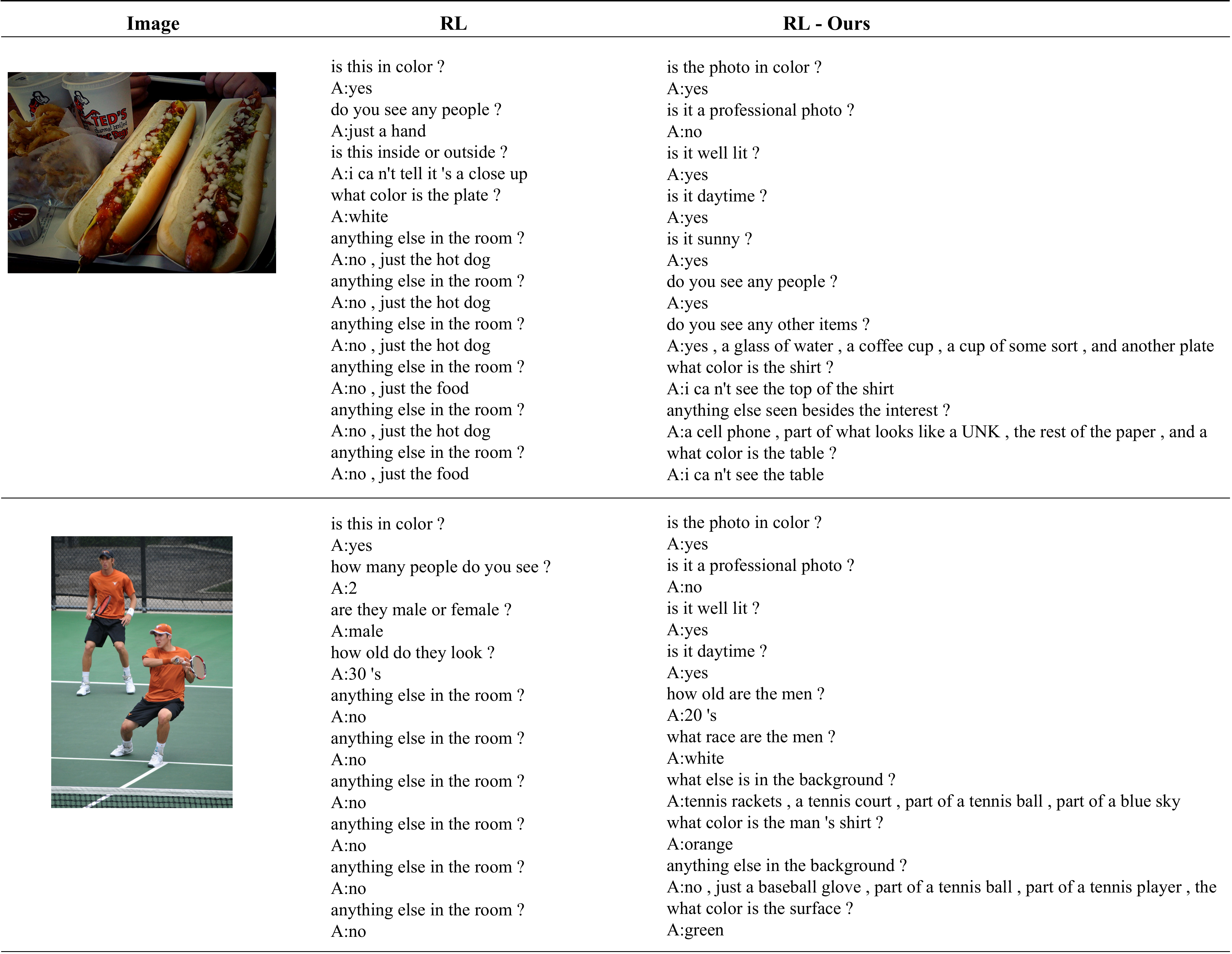}
\end{minipage}
}

\subfigure[]{
\label{fig7:b}
\begin{minipage}[b]{0.9\textwidth}
\includegraphics[width=1\textwidth]{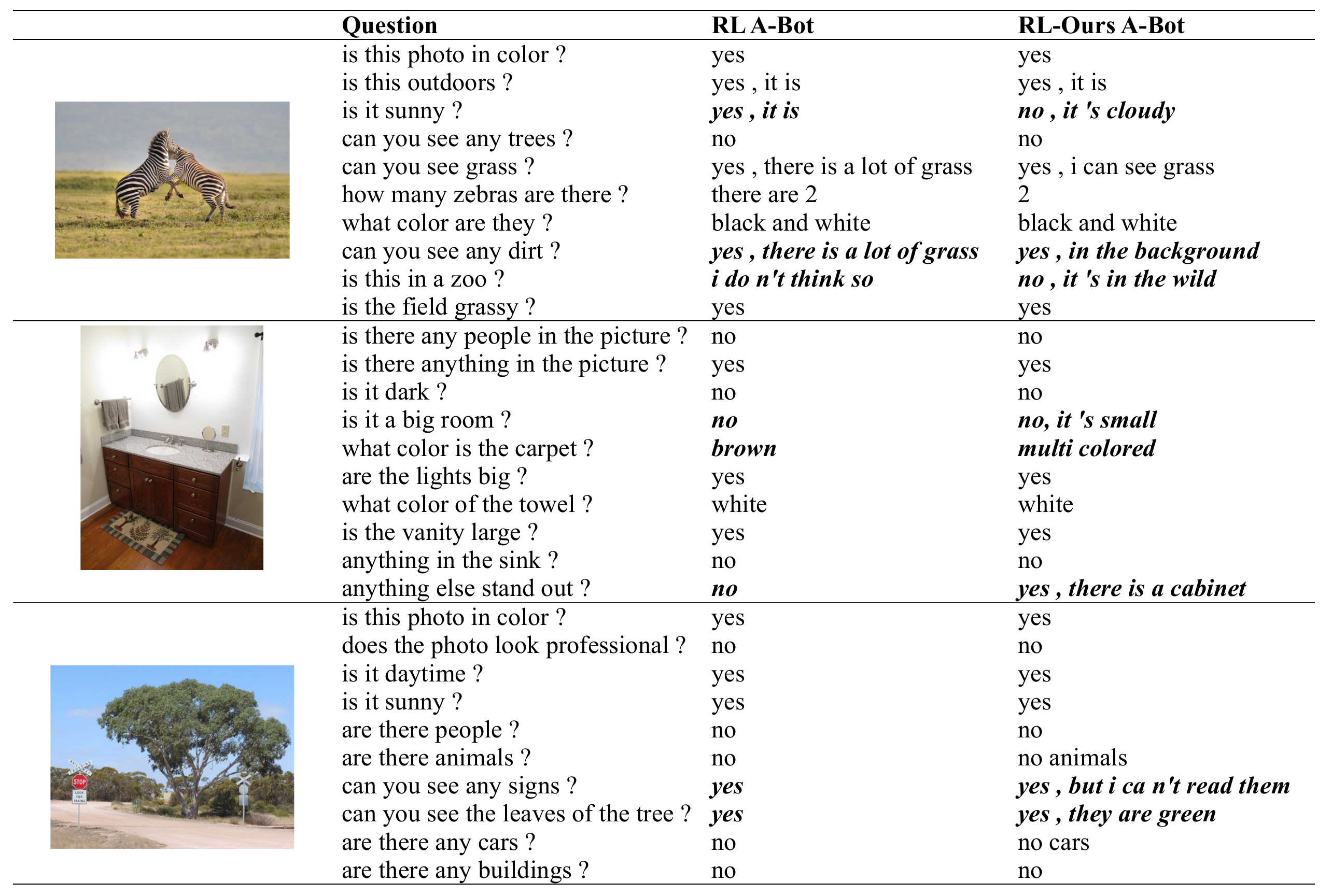}
\end{minipage}
}
\caption{Illustration of A-Bot's ability to generate descriptive responses when (a) answering jointly optimized Q-Bot, (b) answering ground truth questions. Comparing models are: 1)‘RL’: basic agents fine-tuned by RL using the original rewards; 2) RL-Ours: basic agents fine-tuned by RL with two ECS-based rewards in addition.}
\end{figure*}

We find our A-Bot tends to generate descriptive responses that not only directly answer the question but also describe correlated visual information. We give the selected generated dialogue examples in Fig.~\ref{fig7:a}. As can be seen, when is asked questions like \emph{``what else?"}, our A-Bot gives detailed description about visual contents that have not been mentioned yet while the A-Bot optimized by single original reward tediously says \emph{``no"}. As in the top example, our A-Bot constantly adds new information of the undisclosed image by answering \emph{``yes , a glass of water , a coffee cup , a cup of some sort , and another plate"} to \emph{``do you see any other items ?"} and answering \emph{``a cell phone , part of what looks like a UNK , the rest of the paper , and a\ldots"} to \emph{``anything else seen besides the interest ?''}. The descriptive responses help Q-Bot have better knowledge of the image and encourage Q-Bot to generate new questions based on the given detailed answers -- as in the bottom example in Fig.~\ref{fig7:a}, after knowing that there are tennis court in the background, Q-Bot then asks \emph{``what color is the surface ?"}.

In Fig.~\ref{fig7:b}, we give the examples that replace the jointly-trained Q-Bot with the ground-truth questions, to illustrate the adaptability of our A-Bot and make a direct comparison. As can be seen, our A-Bot still performs outstandingly as it tends to give detailed answers that include additional explanations or descriptions moreover the simple yes/no. In the first example, when the question is \emph{``is this in a zoo?"}, our A-Bot answers \emph{``no, it's in the wild"}, which corrects Q-Bot's belief on the undisclosed image, while the basic A-Bot only answers \emph{``i don't think so"}. In the third example, when the question is \emph{``can you see the leaves of the tree?"}, our A-Bot answers \emph{``yes, they are green"} which gives additional information about the related visual contents. Informative and diverse answers like these are more human-like and are helpful to keep users engaged.

\section{Additional Qualitative Analysis on Comparing Methods}
\label{apd:d}
\begin{figure*}[t]
  \centering
  \includegraphics[width=0.95\linewidth]{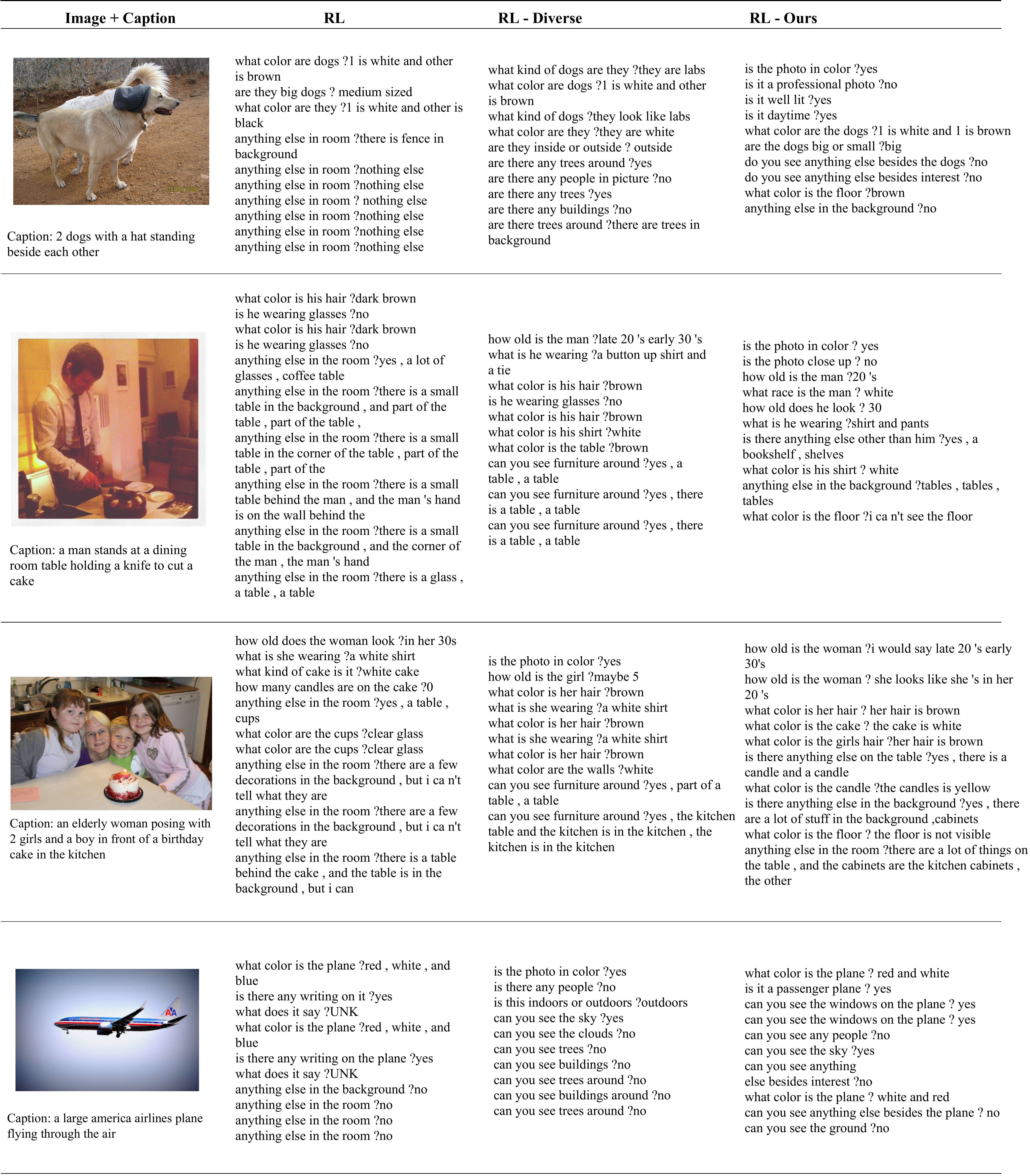}
  \caption{Additional generated dialogue Examples from comparing methods, i.e., 1)‘RL’: basic agents fine-tuned by RL using the original rewards; 2)‘RL-Diverse’: diverse agents fine-tuned by RL using the original rewards; 3) RL-Ours: basic agents fine-tuned by RL with ECS-based rewards in addition.}
  \label{fig8}
\end{figure*}

We give additional generated dialogue examples from comparing methods in Fig.~\ref{fig8}. As what has been analyzed before, our method helps achieve less repetitive and more visual informative dialogues. As can be seen, dialogues generated by \emph{RL} start to repeat at an early stage of the dialogue, dialogues from \emph{RL-Diverse} are less repetitive but tend to repeat at successive rounds, while dialogues generated by \emph{RL-Ours} can always converse on new visual contents. Besides, our dialogues valuably show consistency sometimes. For example, in the 2nd sample, after the A-Bot answers \emph{``shirt and pants''}, the Q-Bot further asks \emph{``What color is his shirt?''}. Also, in the 3rd sample, as the A-Bot mentions \emph{``there is a candle''}, the Q-Bot then asks \emph{``What color is the candle?''}

\section{Analysis on Different Settings of Reward-Coefficients}
\label{apd:e}
We explore the training effects brought by different settings of reward-coefficients, i.e. the coefficients of DR and IR. We show the results in Tab.~\ref{tab:2}, which includes indicators that measure the performances in three-aspects: 1) joint performance of Q-Bot and A-Bot by PMR, i.e. the Percentile Mean Rank in the image-guessing task; 2) Q-Bot performance by Unique Question; 3) A-Bot performance by MRR and NDCG. 

In setting 1, we intuitively assign the reward coefficients to let weighted rewards be in the same order of magnitude. We also conduct experiments with other reward-coefficients settings, as in 2 and 3. We think different coefficient settings lead to different training results on Q-Bot and A-Bot. Setting 1 achieves the most balanced cooperative Q-Bot and A-Bot while other settings may help improve the performances on separate tasks, i.e., image-guessing and Visual Dialogue answering, which are also meaningful.

\begin{table*}[ht]
\setlength{\tabcolsep}{0.35mm}{
\begin{center}
\begin{tabular}{c|ll|cccc}
\hline
settings & \emph{$\alpha_D$}      & \emph{$\alpha_I$}  & $PMR^\text{QA}$ & $Unique question^\text{Q}$   & $MRR^\text{A}$ &  $NDCG^\text{A}$   \\
\hline
1&0.1 & 0.01 & $96.86_\text{$\pm$0.06}$ & $\textbf{8.11}_\text{$\pm$0.08}$ & $49.36_\text{$\pm$0.00}$ & $57.81_\text{$\pm$0.00}$ \\
2&0.05 & 0.01 & $96.57_\text{$\pm$0.08}$ & $6.79_\text{$\pm$0.07}$ & $49.38_\text{$\pm$0.00}$ & \bf $58.00_\text{$\pm$0.00}$ \\
3&0.01 & 0.001 & \bf $97.00_\text{$\pm$0.00}$ & $5.43_\text{$\pm$0.00}$ & \bf $49.42_\text{$\pm$0.00}$ & $57.84_\text{$\pm$0.00}$ \\
\hline
\end{tabular}
\end{center}}
\caption{Performances on VisDial v1.0 val under different reward-coefficients settings. $\alpha_D$ and $\alpha_I$ are coefficients for DR and IR, respectively. The superscripts represent the indicator belongs to which agent or both.}
\label{tab:2}
\end{table*}

\end{document}